\documentclass[letterpaper]{article} % DO NOT CHANGE THIS
\usepackage{aaai23}  % DO NOT CHANGE THIS
\usepackage{times}  % DO NOT CHANGE THIS
\usepackage{helvet}  % DO NOT CHANGE THIS
\usepackage{courier}  % DO NOT CHANGE THIS
\usepackage[hyphens]{url}  % DO NOT CHANGE THIS
\usepackage{graphicx} % DO NOT CHANGE THIS
\usepackage{natbib}  % DO NOT CHANGE THIS AND DO NOT ADD ANY OPTIONS TO IT
\usepackage{caption} % DO NOT CHANGE THIS AND DO NOT ADD ANY OPTIONS TO IT
\usepackage{algorithm}
\usepackage{algorithmic}
\usepackage{amssymb}

\usepackage{amsfonts}
\usepackage{mathtools}
\usepackage{multirow}
\graphicspath{ {figures/} }
\usepackage{subfigure}
\usepackage{color,xcolor}

\usepackage{booktabs}
\usepackage{multirow}
\usepackage{booktabs}
\usepackage{ulem}
\usepackage{bm}
\usepackage{newfloat}
\usepackage{listings}
\DeclareCaptionStyle{ruled}{labelfont=normalfont,labelsep=colon,strut=off} % DO NOT CHANGE THIS
\floatstyle{ruled}
\newfloat{listing}{tb}{lst}{}
\floatname{listing}{Listing}
\title{Conceptual Reinforcement Learning for Language-Conditioned Tasks}
\author {
    Shaohui Peng,\textsuperscript{\rm 1, \rm 2, \rm 3}
    Xing Hu,\textsuperscript{\rm 1}
    Rui Zhang,\textsuperscript{\rm 1, \rm 3}
    Jiaming Guo,\textsuperscript{\rm 1, \rm 2, \rm 3}
    Qi Yi,\textsuperscript{\rm 1, \rm 3, \rm 4} \\
    Ruizhi Chen,\textsuperscript{\rm 2, \rm 5}
    Zidong Du,\textsuperscript{\rm 1, \rm 3}
    Ling Li,\textsuperscript{\rm 2, \rm 5,}
    Qi Guo,\textsuperscript{\rm 1}
    Yunji Chen\textsuperscript{\rm 1, \rm 2, $\dag$}
}
\affiliations{
    \textsuperscript{\rm 1} SKL of Processors, Institute of Computing Technology, CAS\\%, Beijing, China\\
    \textsuperscript{\rm 2} University of Chinese Academy of Sciences \quad%, UCAS, Beijing, China\\
    \textsuperscript{\rm 3} Cambricon Technologies\\
    \textsuperscript{\rm 4} University of Science and Technology of China \quad%, USTC, Hefei, China\\
    \textsuperscript{\rm 5} SKL of Computer Science, Institute of Software, CAS\\%, Beijing, China\\
    {\tt\small  \{pengshaohui18z, huxing, zhangrui, guojiaming18s, duzidong, guoqi, cyj\}@ict.ac.cn, } \\
    {\tt\small  yiqi@mail.ustc.edu.cn, \{ruizhi, liling\}@iscas.ac.cn} \\
}

\usepackage{bibentry}
\begin{document}

\maketitle
{
\let\thefootnote\relax\footnote{$\dag$ Corresponding author.}
} 
\begin{abstract}
Despite the broad application of deep reinforcement learning (RL), transferring and adapting the policy to unseen but similar environments is still a significant challenge.
Recently, the language-conditioned policy is proposed to facilitate policy transfer through learning the joint representation of observation and text that catches the compact and invariant information across environments. 
Existing studies of language-conditioned RL methods often learn the joint representation as a simple latent layer for the given \textit{instances} (episode-specific observation and text), which inevitably includes noisy or irrelevant information and cause spurious correlations that are dependent on instances, thus hurting generalization performance and training efficiency. %\cite{SILG}.
To address this issue, we propose a conceptual reinforcement learning (CRL) framework to learn the \textit{concept-like} joint representation for language-conditioned policy. 
The key insight is that concepts are compact and invariant representations in human cognition through extracting similarities from numerous instances in real-world.
In CRL, we propose a multi-level attention encoder and two mutual information constraints for learning compact and invariant concepts.
Verified in two challenging environments, RTFM and Messenger, CRL significantly improves the training efficiency (up to $70\%$) and generalization ability (up to $30\%$) to the new environment dynamics.
\end{abstract}

\section{Introduction}
Deep reinforcement learning has been successfully applied in various areas such as video games \cite{video1} and robot control \cite{robot1}.
However, it is still a significant challenge to transfer and adapt the policy to unseen but similar environments \cite{surtrans}.
Recently, researchers propose language-conditioned policy that adopts the language as an intermediate information channel to facilitate policy transfer \cite{grounding2, ltrans}.
Based on extra-textual descriptions that specify environment dynamics, the language-conditioned policy can learn the connection between training and testing environments to promote generalization ability.

The key challenge of language-conditioned policy is how to learn the joint representation of observation and textual description.
The joint representation should catch the compact and invariant information about the actual cause of the reward that is shared in various training environments, so that the policy can generalize to unseen environments.
Existing studies of language-conditioned reinforcement learning methods often implicitly learn the joint representation as a simple latent layer for the given \textit{instances} (episode-specific observation and text), which cannot guarantee to catch the invariant factors across different instances \cite{rtfm, messenger, cv, grounding2}.
Such a strategy inevitably includes noisy or irrelevant information and causes spurious correlations that are dependent on instances, thus hurting generalization performance and training efficiency \cite{SILG}.

Inspired by the human activities that extract the similarities across numerous instances in the real-world to form ``concept'' for better understanding the world from the abstract perspective and solving similar tasks efficiently, 
we propose to explicitly learn the \textit{concept-like} representation for language-conditioned policy. Concretely, we mainly consider two important features for such representation, \textit{invariance} and \textit{compactness}, which boost both transfer performance and training efficiency.  
Based on the invariance of concepts, policies can be quickly transferred between similar environments.
Meanwhile, based on the abstraction of concepts, the learned joint representation is more compact, and thus the complexity of policies can be reduced to improve training efficiency.
For example, when playing video games, we can treat various monster instances with different names or appearances as the same concept ``enemy'', and handle them in a similar policy such as attacking them to get drops, thereby adapting to an unseen environment quickly. 
 
Based on the above analyses, we propose a conceptual reinforcement learning (CRL) framework which learns concept-based representation (referred as concepts) for language-conditioned policy to promote sample efficiency and generalization performance in unseen testing environments.
In CRL, we propose a multi-level attention concept encoder and two mutual information constraints for learning concepts. 
Specifically, in the concept encoder, each level of the attention module generates one concept, and the generated concepts participate in subsequent levels of the attention module.
All outputs of the concept encoder are concatenated as the final concept-based representation.
Such a mechanism simulates the multi-step reasoning process and guarantees that the encoder can extract concept-related information from text and observation.
More importantly, we augment two mutual information constraints to guarantee the invariance and compactness of the learned concepts, which is the major difference from latent representation learned implicitly by existing methods.
We minimize mutual information between concepts and entities of observation in different environments, which guides the concepts to catch the shared properties and keep them invariant across different tasks.
Besides, we leverage an information bottleneck between the textual description and concepts to ensure the compactness of the concepts.
Finally, we train the policy upon the generated concept-based representation to drive the learning of concepts by the RL objective.
In conclusion, the concept-based representation learned by CRL has three evident advantages, \textit{invariance}, \textit{compactness}, and \textit{better interpretability}.
First, the invariance significantly promotes the generalization performance of the policy in unseen environments.
Second, the compactness makes the training process of the policy more efficient.
Third, the language-conditioned policy constructed upon concepts has higher interpretability to make the policy easier to employ and understand.

We verify CRL on two challenge environments with textual descriptions, RTFM \cite{rtfm} and Messenger \cite{messenger}, both of which are benchmarks to evaluate the generalization ability of language-conditioned policy to new environment dynamics.
The former focus on multi-step reasoning across observation and textual description, while the latter focus on Out-Of-Distribution (OOD) settings of test environments.
Compared with state-of-the-art methods, the proposed CRL gains significant generalization performance (up to 30\%) and training efficiency promotion (up to 70\%) on both RTFM and Messenger.

\section{Background}
\begin{figure*}[h]
    \centering
    \includegraphics[width=0.85\textwidth]{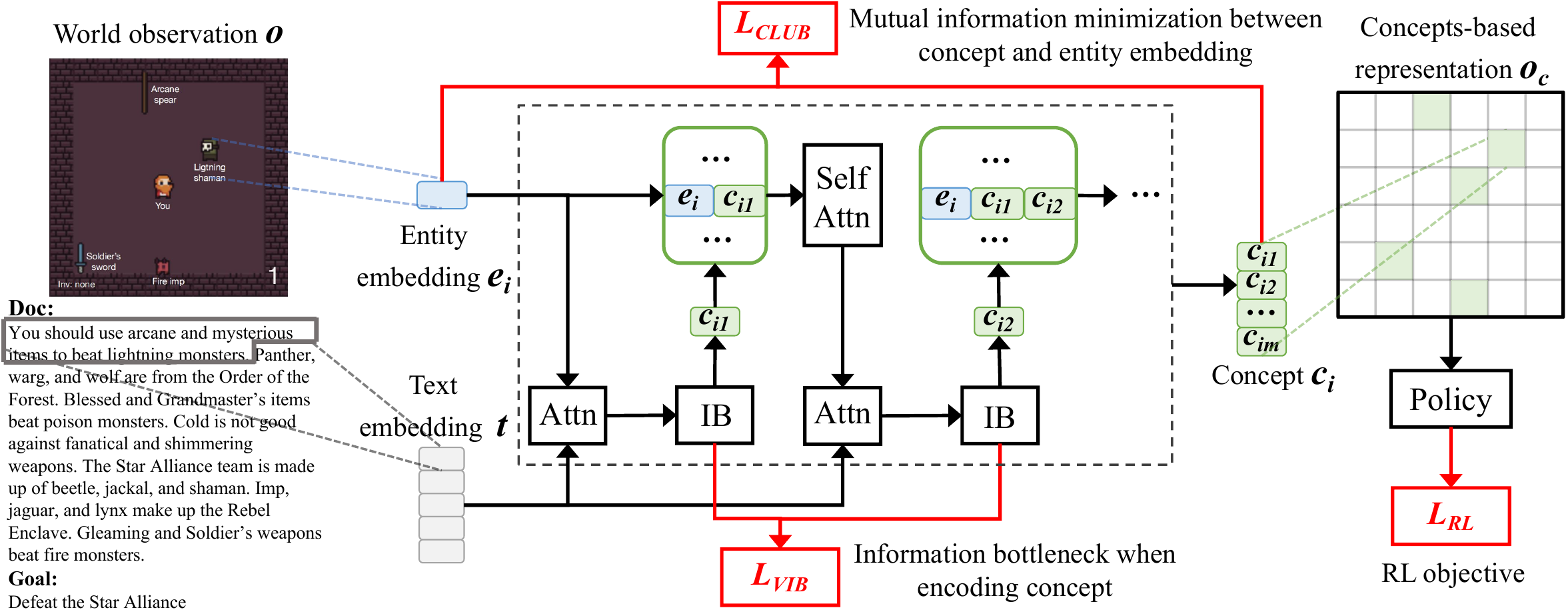}
    \caption{Overview of Conceptual Reinforcement Learning (CRL) framework. Take an example from RTFM. CRL gets inputs including the world observation $o$ that consists of entities and the text $t$. The multi-level concept encoder of CRL leverage entities $e_i$ to conduct multiple attention over the text $t$ to get $e_i$'s concepts $c_i = \{c_{i1}, ..., c_{im}\}$. The previously produced concepts also participate in the subsequent levels of the attention module. Besides, CRL leverages two mutual information constraints to guarantee the two advantages of produced concepts, where $L_{CLUB}$ for invariance and $L_{VIB}$ for compactness. All entities' concepts are combined to form the new concept-based representation $o_c$ for the language-conditioned policy. }
    \label{fig:overview}
\end{figure*}

In this paper, we focused on conceptual reinforcement learning in the language-conditioned policy that applied to the environment with textual description.
The text in such environments specifies the relation of entities, goals, and some environment dynamics, and it is regenerated every episode to initialize different but similar environments.
For example, Read To Fight Monsters (RTFM) is a well-known environment where the agent needs reading and multi-step relation reasoning to understand the relation between tools and monsters.
Compared with other RL environments, the input of the policy in such environments includes two parts, the world observation $o$ and the additional textual description $t$:

\begin{itemize}
    \item \textbf{World observation} $o \in \mathbb{R}^{h\times w\times d}$: $h$ and $w$ represent the height and width of the grid world.
    A $d$-word symbolic name represents the content in the grid.
    Some non-overlapping entities $e \in \mathbb{R}^{n\times d}$ are scatted in the grid world, including the agent and other interactive objects.
    The entities in $o$ except the agent are sampled from an invisible large candidate set and thus may change in every episode.
    In different episodes, the property of the same entities is unfixed and needs to be inferred from the textual description of the current episode. 
    Figure \ref{fig:overview} shows an example of world observation.
    There are five entities in $o$, including the agent, monsters, and tools. 
    \item \textbf{Text} $t \in \mathbb{R}^{num_{sent} \times l_{sent} \times d_{t}}$: 
    The textual description of the environment contains less than $num_{sent}$ sentences, and each sentence consists of less than $l_{sent}$ tokens.
    $d_t$ is the embedding dimension of each token.
    The text keeps fixed in an episode but changes between episodes.
    $t$ describes the goal, the entities' property, and the relation between entities.
    Figure \ref{fig:overview} shows an example of textual description in the RTFM environment.
    ``\texttt{The Star Alliance team is made up of beetle, jackal, and shaman. Defeat the Star Alliance.}'' explains that the entity ``\texttt{shaman}'' is the enemy, and the goal of the agent is to defeat it.
\end{itemize}

The language-conditioned policy $\pi (a|o,t)$ takes world observation $o$ and text $t$ as inputs and outputs an action $a$.
As the text $t$ described some dynamics of the environment and changes every episode, the reward function $r_{\pi}(o,t,a,o^{\prime})$ also takes text $t$ as input.
The target is to find an optimal policy $\pi$ to maximize the accumulated rewards $R_{\pi}(o,t)=E[\sum_{n=0}^{\infty}\gamma^{n}r(o_{t+n},t,a_t,o_{t+n+1})|o_{t}=o]$.
Existing methods implicitly learn the joint representation as a simple latent layer for the given instances (episode-specific observation and text), which cannot guarantee catching the invariant factors across different instances.
Our method (CRL) explicitly learned invariant compact concepts for policy to promote the policy’s generalization performance and training efficiency.

\section{The Conceptual Reinforcement Learning (CRL) Framework}
\begin{figure}[h]
    \centering
    \includegraphics[width=0.45\textwidth]{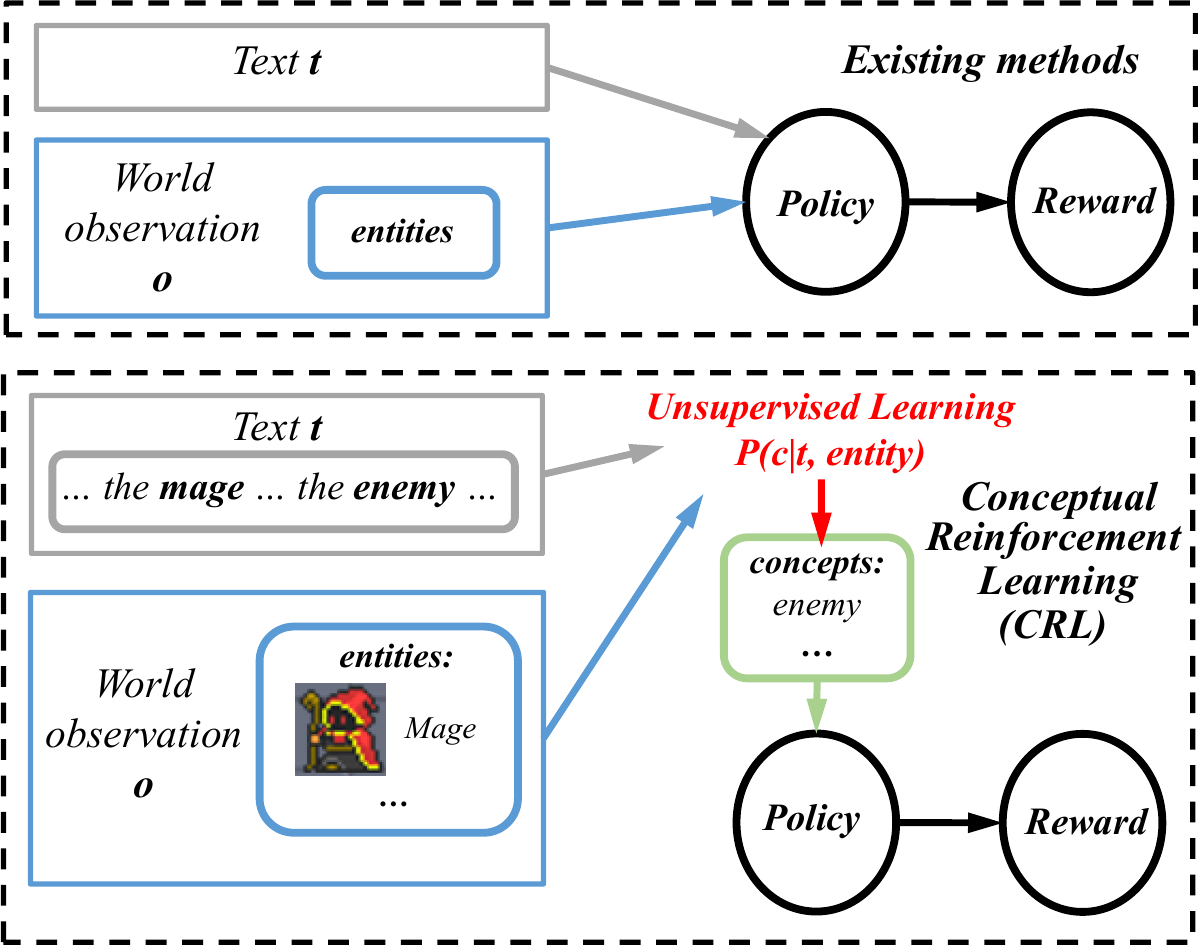}
    \caption{Different modeling of CRL and existing language-conditioned policy. CRL additionally learns the concept encoder $P(c|t, entity)$ compared with existing methods.
    Taking a certain episode as an example, the encoder computes the concept of the entity ``Mage'' as ``enemy''  through reasoning in the text. The concepts are shared properties of different entities in various episodes and are fed into the language-conditioned policy as a joint representation.}
    \label{fig:DGM}
\end{figure}

In this section, we introduce the Conceptual Reinforcement Learning (CRL) framework, aiming to learn the concept-like representation with both advantages of \textit{invariance} and \textit{compactness} across different scenarios for better generalization.
\textit{Invariance} enables the concepts can generalized to unseen similar environments, and \textit{compactness} promotes the training efficiency of the policy.
To achieve these two goals, CRL includes the multi-level attention module as the concept encoder,
which simulates the multi-step reasoning process to extract concept-related information from textual description and observation.
Meanwhile, CRL leverages two mutual information constraints to guarantee the invariance and compactness of the learned concepts.

\subsection{Concept Encoder}
\subsubsection{Motivation}
We are inspired that humans can extract similarities from numerous instances in the real-world to form concepts, which are compact and invariant representations.
We propose introducing concepts as the representation for language-conditioned policy to make the policy be trained efficiently and generalize to similar scenarios.
The analysis of the modeling for the language-conditioned policy of the existing methods and our method that introduce concepts as intermediate representation is as follows.

\textbf{Existing methods}:
Figure \ref{fig:DGM} shows the modeling of existing methods optimizing policy for higher reward relies on all entities and text.
However, entities and the text change every episode and may have distribution shifts on unseen similar scenarios.
For example, the role assignments to the same entity are not overlapped between training and test games in RTFM and Messenger.
The implicitly learned joint representation inevitably catches irrelevant information about entities and further causes spurious correlations.

\textbf{Concept Learning}:
As shown in Figure \ref{fig:DGM}, the CRL framework introduces an explicitly learned representation, called concepts, between inputs and the policy.
Concepts ($c$) are abstractions of entities’ shared properties inferred from the interactive information of text and entities.
For example, there is a role concept in Messenger. Every entity should be a sender, receiver, or decoy regardless of its name or movement patterns.
We model concepts $P(c|t, entity)$ through an unsupervised learning encoder and replace the world observation $o$ as a concept-based world representation consisting of learned concepts.
As a result, the language-conditioned policy only relies on concepts,
which are shared properties between entities that the agent needs to understand and exploit for solving tasks.
Concepts have two evident advantages, \textit{invariance} and \textit{compactness}.
Invariance means that concepts are an invariant representation for entities not only across episodes but also in train or test scenarios, making the policy generalize well in similar unseen environments.
Besides, compared with the large entities candidate set and complex textual description, the concept is a compact representation,
which makes the data utilization of policy training much more efficient.

\subsubsection{Definitions}

In each game episode, the environment offers the world observation $o \in \mathbb{R}^{h\times w\times d}$, which includes $n$ entities $e = \{e_1, \dots, e_n\} \in \mathbb{R}^{n\times d}$, and a textual description $t \in \mathbb{R}^{num_{sent} \times l_{sent} \times d_{t}}$. 
We assume that there are $m$ kinds of invisible concept variables $C = \{C_1, C_2, \dots, C_m\}$ representing the shared properties of entities.
Each entity $e_i$ can be labeled as concatenation of concept values $c_i =\{c_{i1}\, \dots, c_{im}\} \in \mathbb{R}^{m \times d_{c}}$, where $d_{c}$ is the embedding dimension of a single concept.
The concept encoder is to model the distribution of each entity's concepts $p(c_i|e_i,t)$ given the entity $e_i$ and text $t$.
The concepts generated by the concept encoder may not be identical to the ground truth concept since we learn the encoder in an unsupervised way, but the generated concepts have the same advantages as actual concepts, which are invariant and compact.
\subsubsection{Architecture}
The overall design of the CRL framework is replacing $e_i$ in the world observation $o$ as derived concept values $c_{i}$ to construct the concepts-based observation for the policy training. As the most important part of the CRL framework, the 
multi-level attention concept encoder $f_\theta(e_i, t)$ takes in the entity embedding  $e_i \in \mathbb{R}^{d_{e}}$ and text $t$
as input and output concepts $c_{i}$. 
The dotted box in Figure \ref{fig:overview} shows the concept encoder.
The concept encoder contains $m$ Attention modules, and each module outputs a single concept $c_{ij} \in \mathbb{R}^{d_{c}}$. 
These modules produce both basic concepts inferred by raw text information and complex concepts derived from the multi-step reasoning process for the underlying information in the text and basic concepts.

Specifically, the first Attention module could generate concepts inferred by direct information in the text.
For example, in Messenger, different entities have different roles, and the text claims this information explicitly.
``\texttt{the thing that is not able to move is the mage who possesses the enemy that is deadly}''
points out that the role concept of entity ``\texttt{mage}''  is ``\texttt{enemy}''.
In this Attention module, we compute query $q_1$ from the entity embedding $e_i$ and the key $k_1$ and value $v_1$ from the text $t$ as follows:
\begin{equation}
\begin{aligned}
    \bm{q}_1 = & MLP(\bm{e_i}) \\
    \bm{k}_1 = & GRU_{key_1}(\bm{t}) \\
    \bm{v}_1 = & GRU_{val_1}(\bm{t}).
\end{aligned}
\end{equation}
And then, we get the direct information of the concept $c_{i1}$ through the standard Attention module as shown in Figure \ref{fig:overview}.
The following IB module is a Variational Information Bottleneck encoder for the concept $c_{i1}$, aiming to improve the compactness of concepts and will be explained in the next section.

The subsequent Attention modules have an extra Self-Attention model to simulate the multi-step reasoning process to generate complex concepts, considering the acquisition of some complex concepts needs not only raw text information but also the preceding concepts.
For example, in RTFM, ``\texttt{Fire monsters are defeated by fanatical and shimmering weapons.}'' demonstrates the relation between ``\texttt{fire}'' entities and ``\texttt{fanatical}'' entities.
However, to reason whether the entity ``\texttt{fanatical sword}'' is the concept ``\texttt{useful tool}'', we need to know the concept of the ``\texttt{fire}'' entity is an ``\texttt{enemy}'' or not.
So we need to build a multi-level Attention mechanism that concatenates entities with discovered shallow concepts as the query to discover the deep concepts.
As shown in Figure \ref{fig:overview}, the query tensors in the following Attention module are obtained through a Self-Attention mechanism on the concatenation of all entity embeddings and their preceding concepts:
\begin{equation}
\begin{aligned}
     \bm{q}_j = & SelfAttn(Concat(\bm{e_i}, \bm{c_{i1}}, \dots, \bm{c_{i, j-1}}))\\
     \bm{k}_j = & GRU_{key_j}(\bm{t}) \\
     \bm{v}_j = & GRU_{val_j}(\bm{t}),
    where \ \   1<j<=m
\end{aligned}
\end{equation}

Finally, we concatenate all concepts together to get $c_i$, and replace all $e_i, i \in \{0, \dots, n\}$ in the world observation $o$ with their concepts to construct a new concept-based world representation $o_c \in \mathbb{R}^{h \times w \times d_{c}}$.
We use $o_c$ rather than the original world observation $o$ and text $t$ as the representation of the environment state to train the policy $\pi(a|o_c)$. 
\subsection{Mutual Information Constraints}

The model  architecture  of CRL introduced in the last section can obtain an effective concept representation during concept-based policy training. To guarantee the two important features, invariance and compactness, of  produced concepts, we add extra constraints on the concept encoder  $f_\theta$ during policy training. Specifically, we consider the invariance and compactness properties from the view of information theory, detailed as follows. 

\subsubsection{Invariance Constraint}
As the entities change every episode and the properties of the same entity inferred from the text of different episodes may differ, the concepts should only reflect the shared properties across episodes and not contain information about the episode-specific entities.
Therefore, although the concepts are inferred from entities and the text, we need to make the marginal distributions of produced concepts and entities independent to ensure invariance.
As a result, the policy constructed upon the concepts would focus on abstract properties rather than information about episode-specific entities, so it can eliminate spurious correlations.
We minimize the mutual information (MI), a fundamental measure of the dependence between two random variables, to force the learned concept independent of the entity.
Mathematically, the MI between two variables $X$ and $Y$ is: 
\begin{equation}
    I(X;Y)=\mathbb{E}_{p(x,y)}\left[\log \frac{p(x,y)}{p(x)p(y)}\right]
\end{equation}
We adopt CLUB \cite{CLUB} to minimize the upper bound of mutual information $I(e; c)$ between the entities $e$ and its concept $c$ as shown in Figure \ref{fig:overview}.
With $N$ samples $\{e_i,c_i\}_{i=1 \to N}$, the upper bound of MI estimated by CLUB is: 
\begin{equation}
\begin{aligned}
    I_{CLUB}(e;c)&=  \\
    \frac{1}{N^2}\sum\limits^N\limits_{i=1}\sum\limits^N\limits_{j=1}&\left[\log q(c_i|e_i)-\log q(c_j|e_i)\right],
\end{aligned}
\end{equation}
where $q(c_i|e_i)$ is a multi-layer MLP predictor.% trained through supervised learning.
We can minimize mutual information between entities and their concepts through the following extra loss:
\begin{equation}
L_{CLUB}(\theta) = \sum\limits^{n}_{i=1} I_{CLUB}(c_i; e_i) = \sum\limits^{n}_{i=1} I_{CLUB}(f_\theta(e_i, t); e_i),    
\end{equation}
where $n$ is the number of entities in the world observation $o$.

\subsubsection{Compactness Constraint}
The RL training process implicitly ensures the %completeness
effectiveness
of concepts since the agent solves tasks using concept-based observation. 
We propose an additional compactness constraint to eliminate noisy or irrelevant information during concept encoding, so that only necessary information for the task is included in the concepts. 
We adopt an information bottleneck (IB) to compress the concepts, a technique in the information theory to compress the noise information from the input variable $X$ and retain the representative information for the output variable $Y$ in the middle variable $Z$. The mutual information objective of IB is as follows:
\begin{equation}
    \underset{p(z|x)}{min} \ I(X;Z)-\beta I(Z;Y)
    \label{IB}
\end{equation}
In our work, 
as the target of retaining useful information is implicitly done through the RL objective,
we only utilize the first term in equation \ref{IB} to compress irrelevant information when encoding text $t$ to the concept $c$.
We adopt Deep Variational Information Bottleneck (Deep VIB) to minimize the upper bound of $I(c; t)$ as shown in Figure \ref{fig:overview}.
In practice, we implement the encoder head (IB in Figure \ref{fig:overview}) of $f_\theta$ with the form of $\mathcal{N} (c_{ij}|\mu^{d_{c}}(t_{ij}^\prime), \sigma^{d_{c}}(t_{ij}^\prime))$, representing the Gaussian distribution of the $j$-th concept of $e_i$.
And $t_{ij}^\prime$ is the output of the $j$-th Attention module containing the information of the concept, $\mu^{d_{c}}$ and $\sigma^{d_{c}}$ are MLP layers to output distribution parameters with $d_{c}$ dimensions.
Then we can sample a noise $\epsilon$ from $\mathcal{N}(0,1)$ to compute the concept $c_{ij}$ using the reparameterization trick.
We minimize the Kullback-Leibler divergence between encoder head $p_{\theta}(c_{ij}|t_{ij}^{\prime})$ and Normal distribution $\mathcal{N}(0,1)$ to compress the noise and irrelevant information:
\begin{equation}
    L_{VIB}(\theta)=\frac{1}{N}\sum\limits^{N}\sum\limits_{i=1}^{n}\sum\limits_{j=1}^{m}\mathbb{E}_{\epsilon \sim p(\epsilon)}\left[KL(p_\theta(c_{ij}|t_{ij}^{\prime}),\mathcal{N}(0,1))\right],
\end{equation}
where $N$ is the number of training samples.

In summary, to keep the invariance and compactness of the concept, we add extra mutual information constraints $L_{CLUB}$ and $L_{VIB}$ on the concept encoder when training policy.
Concretely, these two constraints can be easily integrated into the standard RL method.
The new objective is: 
\begin{equation}
   %L(\theta) = \mathbb{E}_t \left[ \log \pi_\theta(a_t|s_t) A_t \right] + \alpha_{1} L_{club}(\theta) +\alpha_{2} L_{vib}(\theta),
   L_{CRL}(\theta) = L_{RL}(\theta) + \alpha_{1} L_{CLUB}(\theta) +\alpha_{2} L_{VIB}(\theta),
\end{equation}
where $L_{RL}(\theta)$ is the original RL objective, and coefficients $\alpha_{1}, \alpha_{2}$ are hyperparameters. (Details are in Appendix B). 

\section{Experimental Setup}
\subsection{Environments}
To verify the performance of CRL, we evaluate the framework on two challenging benchmarks, RTFM and Messenger.
\textbf{RTFM} \cite{rtfm} is a game that the agent needs to \textbf{R}ead \textbf{T}ext to obtain the correct tool in the grid world and then \textbf{F}ight the correct \textbf{M}onster. The key challenge of RTFM is that the agent needs multi-step reasoning to interpret the text. RTFM has a train set of environment dynamics (including entities and role assignments) and an independently identically distribution (i.i.d.) held-out test set. The environment also has four kinds of complexity settings, including base $6\times 6$ grids setting (\texttt{$6\times 6$}), describing entities in the form of many-to-one group assignments to make disambiguation more difficult (\texttt{group}), allowing moving monsters that hunt down the player (\texttt{dyna}) and using natural language descriptions (\texttt{nl}). Because of the high complexity, the environment also offers four curriculum stages. 
\textbf{Messenger} \cite{messenger} is also a grid environment ($10\times 10$). The agent must extract the role information of the entities from the text manual to acquire a message from the sender and deliver it to the receiver. Unlike RTFM, Messenger's key challenge is the out-of-distribution (OOD) problem. The entities and role assignments are no-repeat and have distribution shifts between train and test scenarios, which may lead to spurious correlations. The Messenger offers three difficulty stages, including only message acquiring or delivering (\texttt{S1}), both acquiring and delivering (\texttt{S2}), and adding decoy entities and irrelevant descriptions (\texttt{S3}).

The details of the environment and the CRL implementation are shown in Appendix A and B.  

\subsection{Baselines}
We compare CRL with two state-of-art methods, \textbf{txt2$\bm{\pi}$} on RTFM and \textbf{EMMA} on Messenger, respectively.

\textbf{txt2$\bm{\pi}$} \cite{rtfm} exploits the prior knowledge of the observation structure in RTFM, which builds the representation by capturing the three-way interactions between the goal description, the environment dynamic description, and the world observation.
txt2$\pi$ outperforms the language-conditioned CNNs and the Feature-wise linear modulation (FiLM) on RTFM.
\textbf{EMMA (Entity Mapper with Multi-modal Attention)} \cite{messenger} uses the entity-conditioned attention module to select relevant information in the text to generate the representation.
It achieves the best performance on Messenger compared with other baselines.

\section{Results}
We first compare the performance of CRL and the state-of-the-art in RTFM and Messenger to show its effectiveness. 
Then, we conduct ablation studies to show the influential impact of the mutual information constraints and pre-defined concepts number during concept generation. 
At last, we visualize the concept embeddings in both environments to intuitively show what CRL learns.

\begin{table*}[]
\centering
\scalebox{0.8}{
\begin{tabular}{@{}lcccccccccc@{}}
\toprule
\multirow{2}{*}{\textbf{\begin{tabular}[c]{@{}l@{}}Transfer\\ from\end{tabular}}} &
  \multirow{2}{*}{\textbf{Method}} &
  \multicolumn{8}{c}{\textbf{Transfer to}} &
  \multirow{2}{*}{\textbf{\begin{tabular}[c]{@{}l@{}}Training\\ Steps\end{tabular}}} \\ \cmidrule(l){3-10} 
 &
   &
  $6\times 6$ &
  \begin{tabular}[c]{@{}c@{}}$6\times 6$\\ \texttt{dyna}\end{tabular} &
  \begin{tabular}[c]{@{}c@{}}$6\times 6$\\ \texttt{groups}\end{tabular} &
  \begin{tabular}[c]{@{}c@{}}$6\times 6$\\ \texttt{nl}\end{tabular} &
  \begin{tabular}[c]{@{}c@{}}$6\times 6$\\ \texttt{dyna}\\ \texttt{groups}\end{tabular} &
  \begin{tabular}[c]{@{}c@{}}$6\times 6$\\ \texttt{groups}\\ \texttt{nl}\end{tabular} &
  \begin{tabular}[c]{@{}c@{}}$6\times 6$\\ \texttt{dyna}\\ \texttt{nl}\end{tabular} &
  \begin{tabular}[c]{@{}c@{}}$6\times 6$\\ \texttt{dyna}\\ \texttt{groups}\\ \texttt{nl}\end{tabular} \\ \midrule
\multirow{2}{*}{random} &
  txt2$\pi$ &
  $84\pm 20$ &
  $26\pm 7$ &
  $25\pm 3$ &
  $45\pm 6$ &
  $23\pm 2$ &
  $25\pm 3$ &
  $23\pm 2$ &
  $23\pm 2$ &
  $100M$ \\ %\cmidrule(l){2-11} 
 &
  CRL &
  \textbf{$\bm{93\pm 3}$} &
  $36\pm 10$ &
  $33\pm 10$ &
  $38\pm 4$ &
  $17\pm 2$ &
  $20\pm 2$ &
  $17\pm 3$ &
  $16\pm 3$ &
  $\bm{30M(\downarrow 70\%})$ \\ \midrule
\multirow{2}{*}{$6 \times 6$} &
  txt2$\pi$ &
   &
  \textbf{$\bm{85\pm 9}$} &
  $82\pm 19$ &
  $78\pm 24$ &
  $64\pm 12$ &
  $52\pm 13$ &
  $53\pm 18$ &
  $40\pm 8$ &
  $50M$ \\ %\cmidrule(l){2-11} 
 &
  CRL &
   &
  \textbf{$\bm{89\pm 9}$} &
  \textbf{$\bm{96\pm 3}$} &
  \textbf{$\bm{97\pm 2}$} &
  \textbf{$\bm{85\pm 1}$} &
  \textbf{$\bm{87\pm 2}$} &
  \textbf{$\bm{85\pm 4}$} &
  \textbf{$\bm{86\pm 3}$} &
  $\bm{30M(\downarrow 40\%})$ \\ \bottomrule
\end{tabular}}
\caption{RTFM results in different settings. The results of ``transfer from random'' means training from random initialization. Each cell show final mean and standard deviation of win rate of 5 random initialized model.}
\label{tb:rtfm}
\end{table*}
\begin{table}[t]
\centering
\scalebox{0.8}{
\begin{tabular}{@{}clllc@{}}
\toprule
\multicolumn{1}{l}{} & \textbf{Method} & Train & Test (OOD) & \multicolumn{1}{l}{\textbf{\begin{tabular}[c]{@{}l@{}}Training\\ Steps\end{tabular}}} \\ \midrule
\multirow{2}{*}{\textbf{S1}} & EMMA & $\bm{98\pm 2.1}$ & $\bm{85\pm 1.4}$ & 30M \\ %\cmidrule(l){2-5} 
                                  & CRL   & $\bm{98\pm 1}$   & $\bm{88\pm 2.5}$ & 30M \\ \midrule
\multirow{2}{*}{\textbf{S2}} & EMMA & $\bm{96\pm 2}$   & $45\pm 12$       & 30M \\ %\cmidrule(l){2-5} 
                                  & CRL   & $\bm{98\pm 2}$   & $\bm{76\pm 5}$   & 30M \\ \midrule
\multirow{2}{*}{\textbf{S3}} & EMMA & $19\pm 2.9$      & $10\pm 0.8$      & 30M \\ %\cmidrule(l){2-5} 
                                  & CRL   & $\bm{43\pm 5}$   & $\bm{32\pm 1.9}$ & 30M \\ \bottomrule
\end{tabular}
}
\caption{Messenger results in different stages. Train scenarios are training from scratch, and then zero-shot transfer to Test scenarios. Each cell shows the mean and standard deviation of the win rate and gets from 5 random seeds.}
\label{tb:mes}
\end{table}
\subsection{Transfer in RTFM}
We compare the transfer performance of CRL and txt2$\pi$ in the RTFM environment, as shown in Table \ref{tb:rtfm}. CRL outperforms txt2$\pi$ in terms of both the transfer performance and training efficiency.
\emph{1). Transfer performance:}
CRL outperforms txt2$\pi$ on the base $6 \times 6$ environment and all variant environments when transferred from the base $6\times 6$. The results show that after training on the $6 \times 6$ environment, CRL can successfully transfer the learned policy to all variant environments and get up to $46\%$ promotion on win rate. In comparison, txt2$\pi$ needs four curriculum stages to achieve a similar win rate (details in Appendix A). This is because CRL learned invariant concepts in the base environment that can be reused in other environments quickly. 
\emph{2). Training efficiency:}
Besides, CRL also gets large training efficiency promotion. CRL saves $70\%$ training steps (100 million to 30 million) on $6 \times 6$ and $40\%$ on other transfer environments (50 million to 30 million). The learning curves in Figure \ref{fig:rtfm_abl} show the efficiency comparison intuitively. The reason for efficiency promotion is the compactness of the concept. In conclusion, benefiting from the invariance and compactness of the concept, the policy learned through CRL can transfer to similar new environments efficiently.
\subsection{Zero-shot generalization in Messenger}
Different from the i.i.d setting of RTFM, there are out-of-distribution (OOD) problems between the training and test environments in Messenger.
We compare the training performance and zero-shot generalization performance on unseen test games.
\emph{1). Zero-short generalization performance:} Table \ref{tb:mes} shows that CRL outperforms EMMA on the more challenging stages (S2 and S3), including the higher training win rate (up to 20\% in S3) and lower generalization gap (up to 30\% in S2). The results show that the invariant concepts can mitigate spurious correlations in OOD generalization scenarios.
\emph{2). Training efficiency:} CRL also promotes the training efficiency on Messenger (see Figure \ref{fig:mes_abl}), but the lift is less than that in RTFM.% since Messenger is relatively simple.
\subsection{Ablation study}
\begin{figure}
\centering
\subfigure[RTFM-base]{\includegraphics[width=0.23\textwidth]{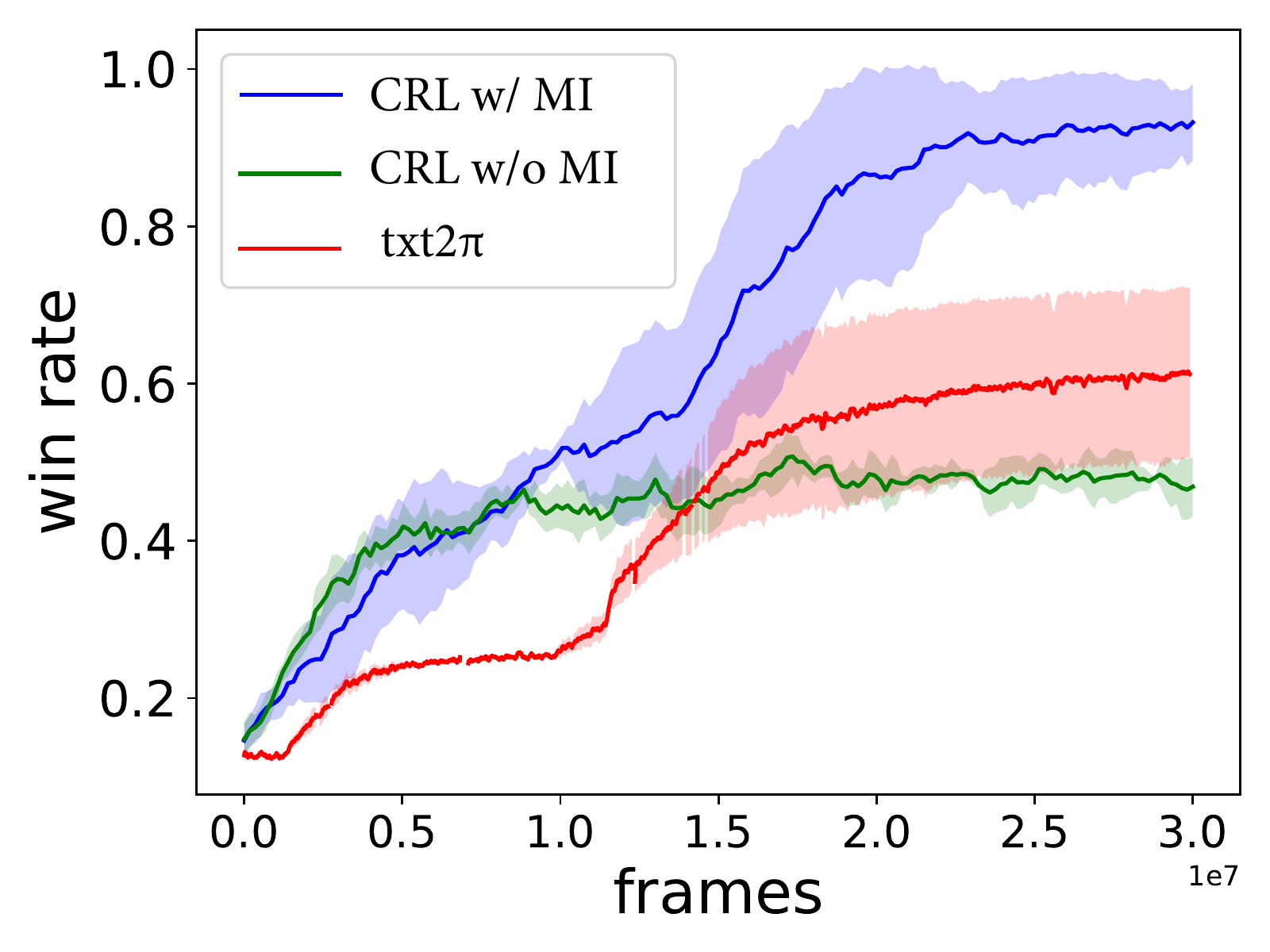}}
\subfigure[RTFM-final]{\includegraphics[width=0.23\textwidth]{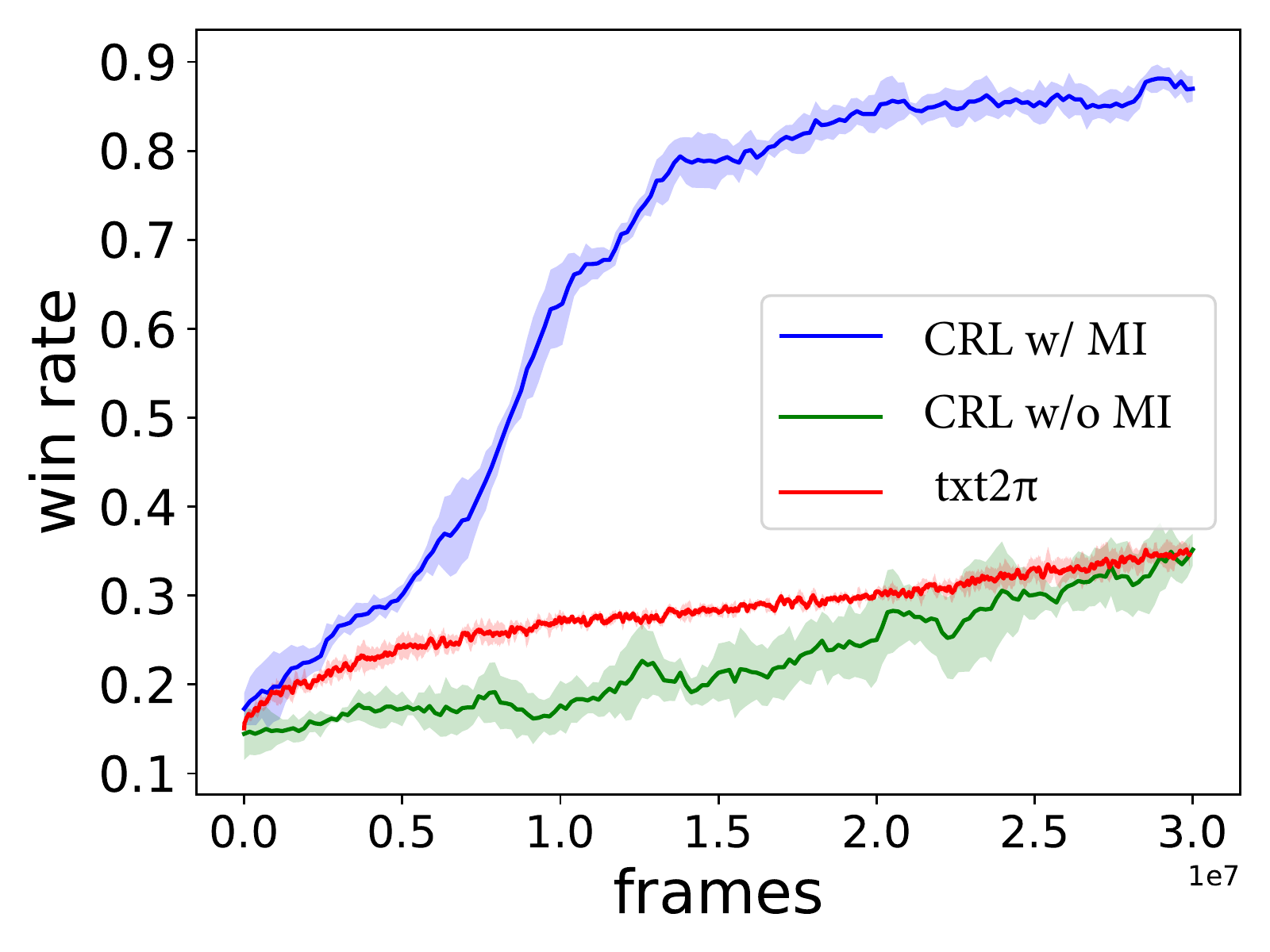}}
\caption{Ablation results in RTFM}
\label{fig:rtfm_abl}
\end{figure}

\begin{figure}
\centering
\subfigure[Messenger-S2]{\includegraphics[width=0.23\textwidth]{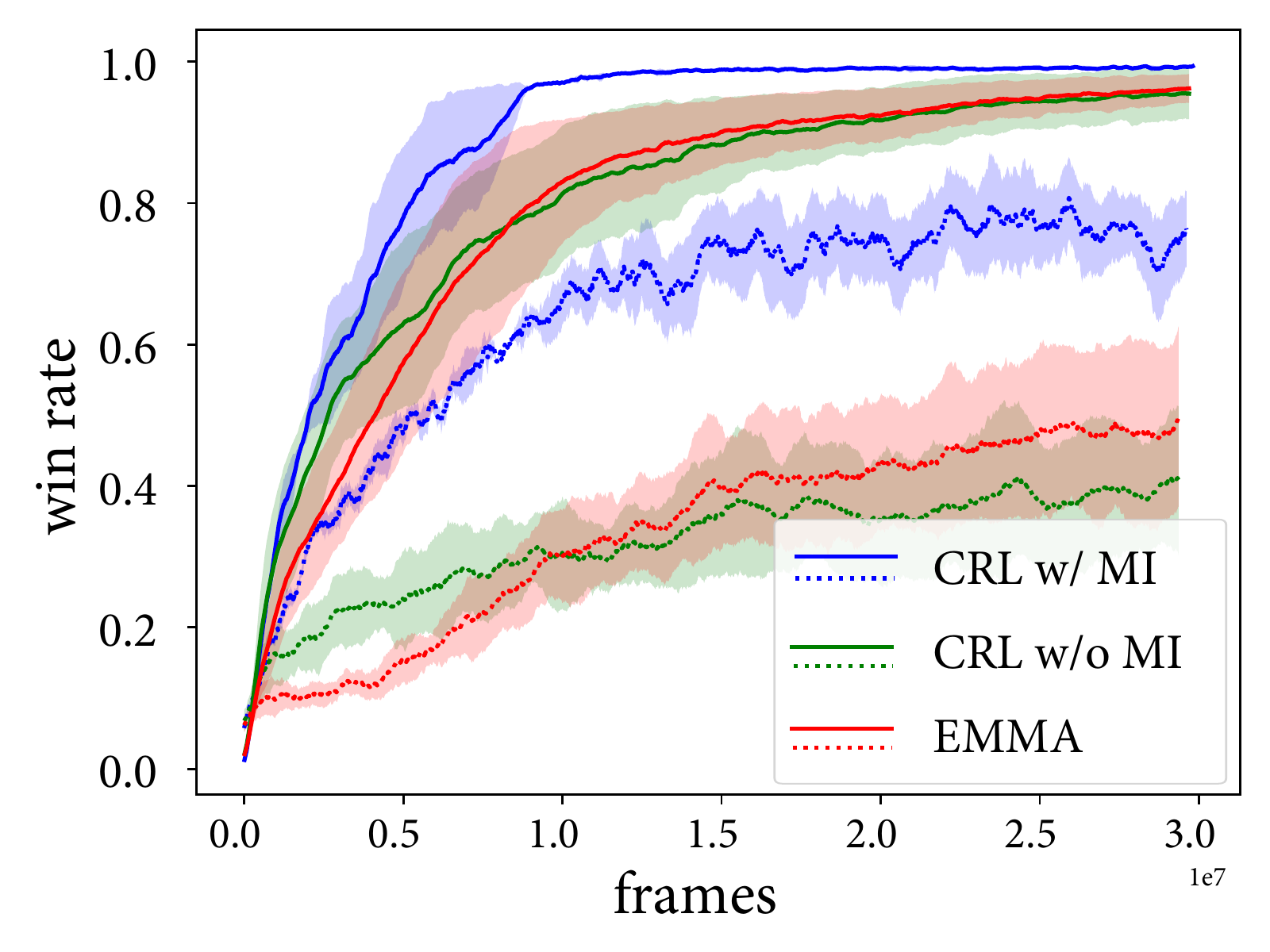}}
\subfigure[Messenger-S3]{\includegraphics[width=0.23\textwidth]{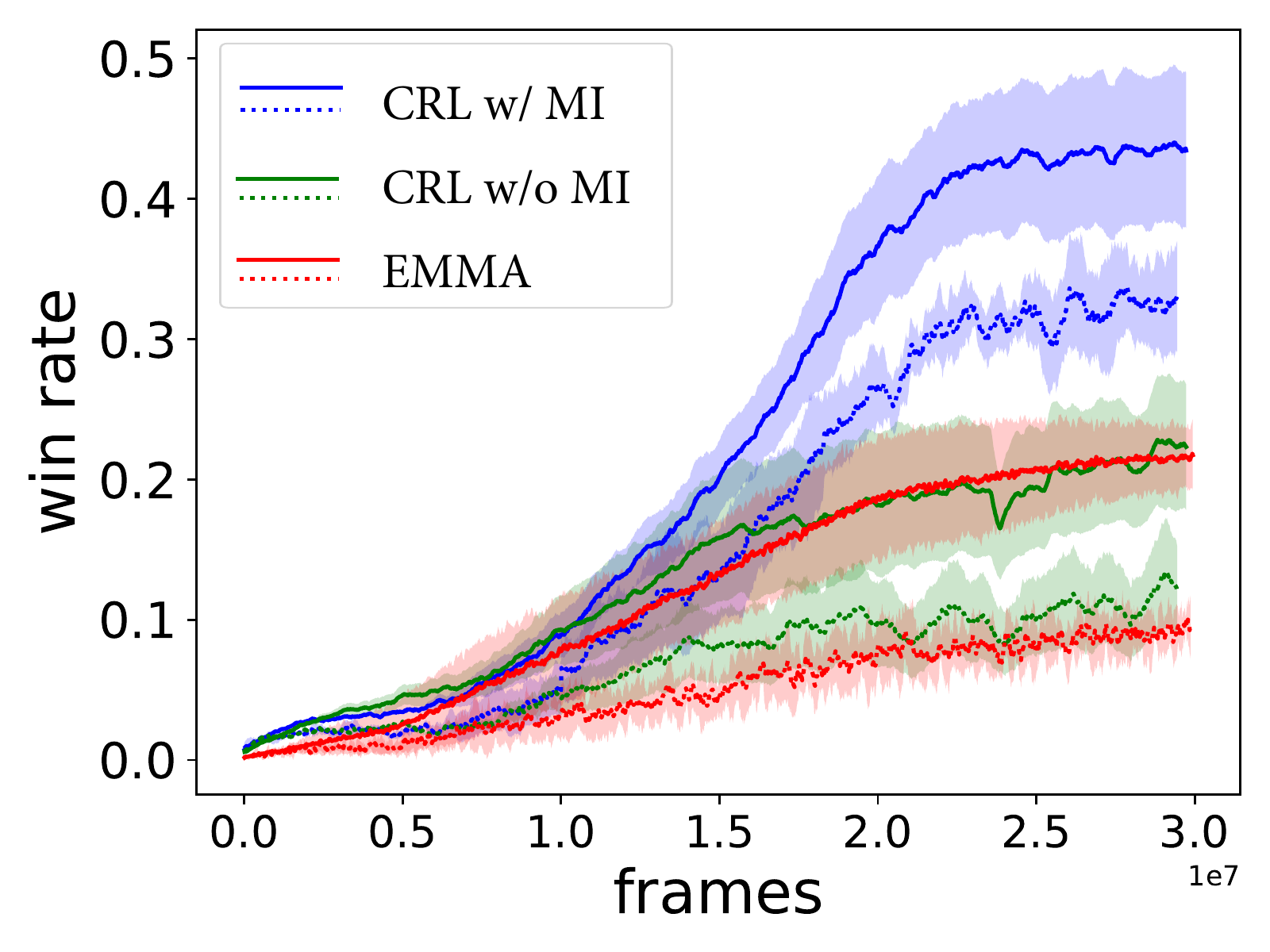}}
\caption{Ablation results in Messenger. The solid and dotted curves represent training and test results.}
\label{fig:mes_abl}
\end{figure}

\begin{figure}
\centering
\subfigure[RTFM-base]{\includegraphics[width=0.23\textwidth]{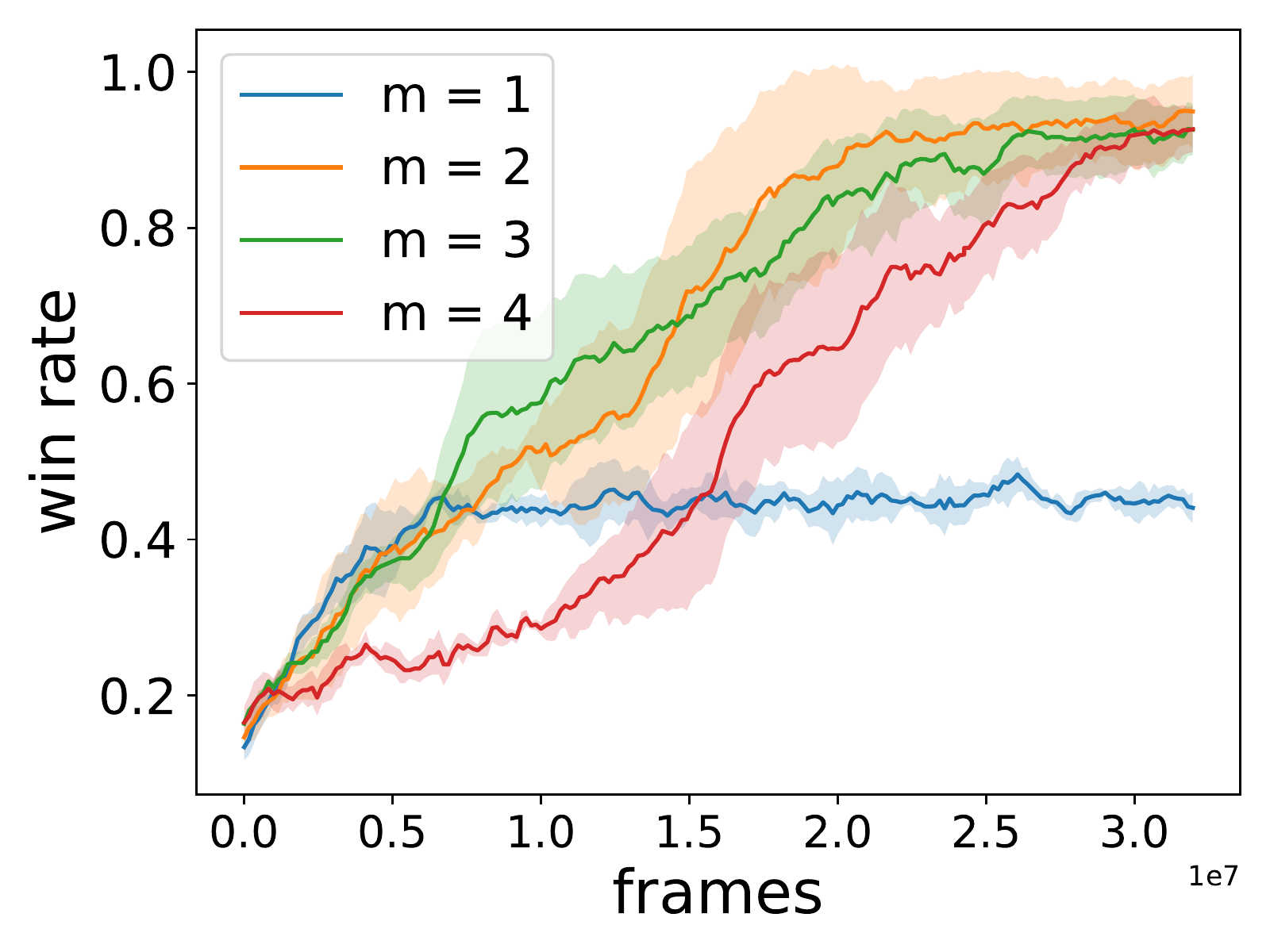}}
\subfigure[RTFM-final]{\includegraphics[width=0.23\textwidth]{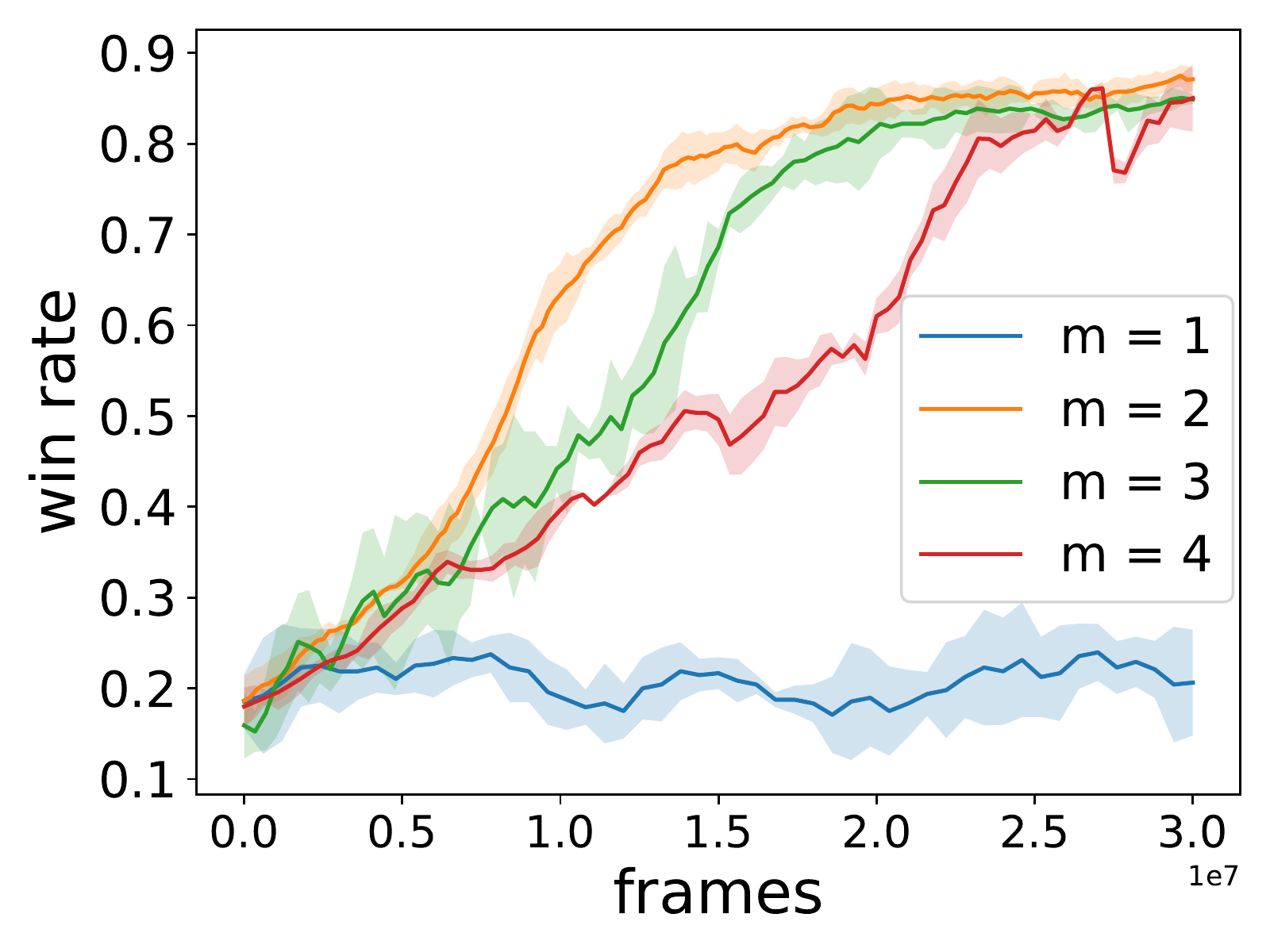}}
\caption{Ablation study of pre-defined concept number (m)}
\label{fig:rtfm_abl_m}
\end{figure}

\paragraph{\textbf{Mutual information (MI) constraints:}}
To show the significant effect of the MI constraints when learning concepts, we verified the performance of the ablated variant CRL w/o MI.
CRL w/o MI shares the same architecture with CRL, but without the MI constraints.
Figures \ref{fig:rtfm_abl} and \ref{fig:mes_abl} show the comparison of learning curves.
The results show that the performance of CRL with MI constraints outperforms the previous method apparently, especially in generalization ability.
In contrast, the ablation variant degenerates to other baselines that implicitly learn the latent joint representation.
The ablation experiments proved the importance of MI constraints when learning compact and invariant concepts (The ablation study of separate MI constraints refers to Appendix C).
\paragraph{\textbf{Pre-defined concept number:}}
We also investigate the influence of the pre-defined concept number $m$ on CRL. Considering the environment RTFM, whose ground truth $m$ is 2, we test $m=1$ to $m=4$, and compare the performance in RTFM-base and transfer performance on the hardest RTFM-final respectively. Figure \ref{fig:rtfm_abl_m} shows that when the pre-defined $m$ is larger than the ground truth value, the efficiency is slightly influenced, while the asymptotic performance is stable. The reason is that the encoder model has more parameters and becomes harder to train, but the MI constraints can guarantee the learned concepts are compact and invariant. When the pre-defined $m$ is smaller than the ground truth, which makes the encoder cannot exact concept-related information, the performance is affected obviously. The above results inspired the user to set a larger concept number $m$.
\subsection{Visual results of the concept}
\begin{figure}[t]
\centering
\subfigure[Colored by concept labels]{\includegraphics[width=0.15\textwidth]{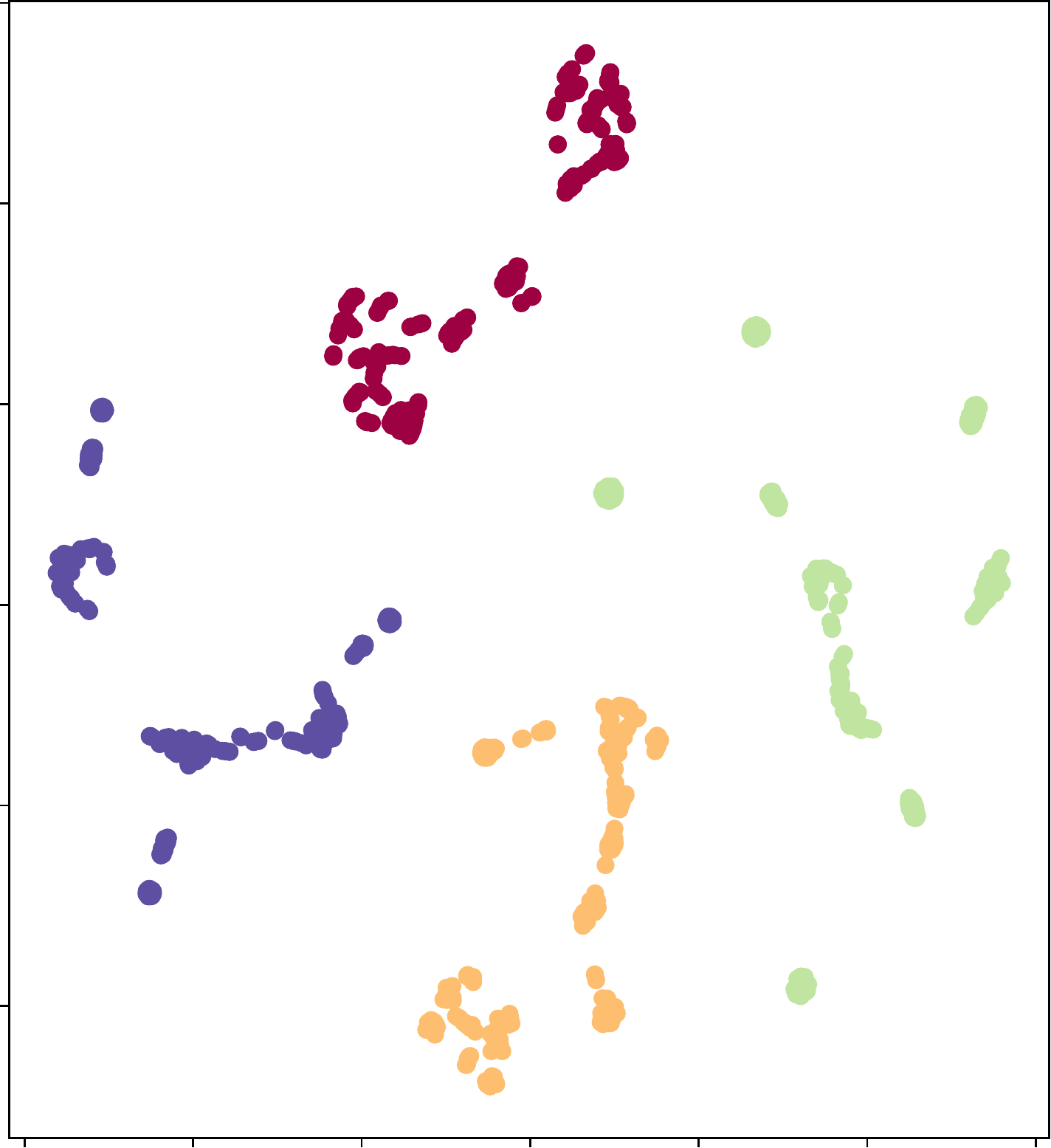}}
\subfigure[Colored by entity ID labels]{\includegraphics[width=0.15\textwidth]{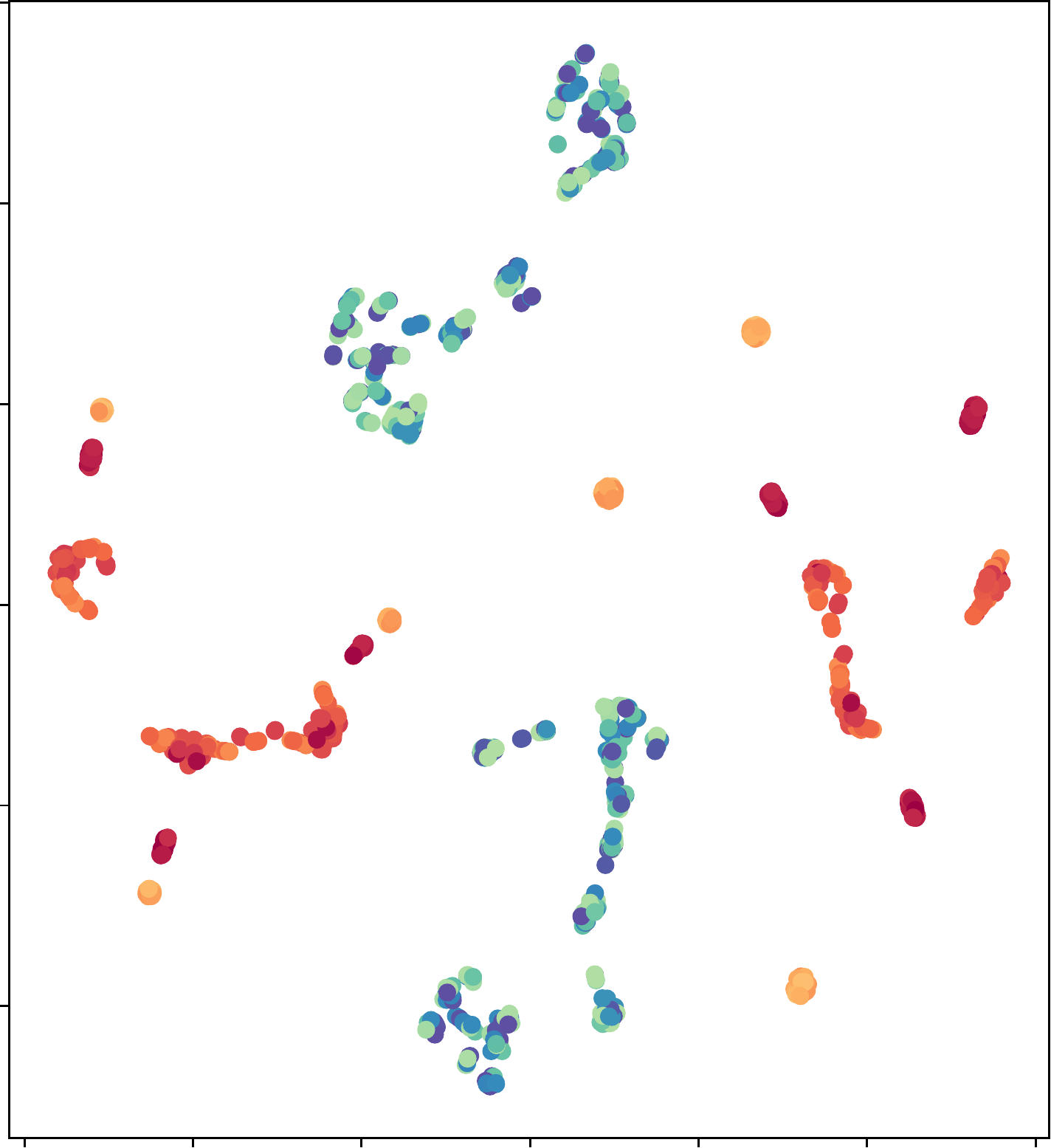}}
\caption{Visual results of \textbf{concept vectors} in RTFM}
\label{fig:v_r_c}
\end{figure}

To intuitively show concepts learned by CRL, we represent the TSNE (t-Distributed Stochastic Neighbor Embedding) results of learned concept vectors in Figures \ref{fig:v_r_c}.% and \ref{fig:v_m_c}.
Specifically, we randomly initialize the environments and save the concept vectors of the entities generated by CRL, then compute the TSNE results of 1000 concept vectors.
RTFM has two concepts from the human view, and each has two values: goal/decoy monster and useful/useless tools.
Figure \ref{fig:v_r_c} shows the TSNE results colored by concept labels (unavailable for the policy) and entity IDs, respectively.
Results show that the learned concept vectors separate into 4 clusters corresponding to four concept values.
Furthermore, the same concept of different entities aggregates to the same cluster.
The phenomenon implies that using entity straightforwardly as the observation for policy training may introduce irrelevant information and cause spurious correlations, while produced concepts by CRL can keep invariant across different entities of various scenarios to address this issue.
% Figure \ref{fig:v_m_c} shows similar results in Messenger, which has one role concept with three values, sender, receiver, and decoy.
% The concepts also form into 3 clusters despite different entities.
% \begin{figure}[t]
% \centering
% \subfigure[Colored by concept labels]{\includegraphics[width=0.15\textwidth]{mes_tsne_cc.pdf}}
% \subfigure[Colored by entity ID labels]{\includegraphics[width=0.15\textwidth]{mes_tsne_ce.pdf}}
% \caption{Visual results of \textbf{concept vectors} in Messenger}
% \label{fig:v_m_c}
% \end{figure}
\section{Related works}
\textbf{Language-conditioned Reinforcement Learning}:
Recently, there have been many works combining reinforcement learning with language and learning language-conditioned policy.
The most related field is \textbf{reading to act}, for example, RTFM \cite{rtfm} and Messenger \cite{messenger}.
The agent is additionally provided information of the environment dynamics through text.
Previous methods often implicitly learn the joint representation of observation and text when optimizing the policy \cite{messenger, grounding2, rtfm, SILG, cv}.
The implicit joint representation will inevitably include noisy or irrelevant information and cause spurious correlations.
Our method explicitly learns the invariant compact representation to address this issue.

Besides reading to act, there are two kinds of scenarios, instruction following and text game.
\textbf{Instruction following} means that agents need to solve different tasks specified by high-level instructions in the environment \cite{ifenv1, ifenv2, ifenv3}.
Different instruction indicates different reward functions in the same environment dynamic, which limits the generalization ability of the agent across different dynamics.
Traditional methods require prior knowledge to model the relation between instructions and observations \cite{if1}.
More recently, some works directly embed both the observation and instruction as the joint input of the policy and train the policy under the reinforcement learning paradigm \cite{if2, if3, if4}.
\textbf{Text game} \cite{tgsurvey} environments are transformed from Interactive Fiction Games where the agent must interact with the environment through textual observation and action, such as TextWorld \cite{textworld} and Jericho \cite{Jericho}.
The language-conditioned RL methods in the text game focus on exploration \cite{tgexp1, tgexp2}, action space reduction \cite{tgaction1, tgaction2, tgaction3}, and reward shaping \cite{tgreward1, tgreward2}.

\textbf{Invariant Representation Learning}
Learning invariant representation that can generalize across environments is critical to the application of RL algorithms \cite{invrep1, invrep2}.
Some works leverage inductive biases, for example, object-oriented architecture \cite{invrep_induc1,yi2022objectcategory} and disentangled representation \cite{invrep_induc2,peng2022causalitydriven}.
Some methods want to eliminate misleading information like temporal information \cite{invrep_info1,jiaming} or background \cite{invrep_info2}.
Other methods define the metric of invariance for optimizing the representation, like Bisimulation metrics \cite{invrep_m1} and policy similarity \cite{invrep_m2}.
The above invariant representation Learning methods cannot work for language-condition policy.
Our method combines the attention-based reasoning module and mutual information constraints to learn an invariant and compact representation for language-conditioned policy.

\section{Conclusion}
The language-conditioned policy is proposed to facilitate policy transfer through learning the joint representation of observation and text that catches the compact and invariant information across various environments.
We propose a conceptual reinforcement learning (CRL) framework to learn the joint representation with both advantages of \textit{invariance} and \textit{compactness} for language-conditioned policy.
Verified in two challenging environments, RTFM and Messenger, CRL significantly improves the training efficiency and generalization ability to the new environment dynamics.
As concept-like representation is valuable for general policy, we hope to broaden the application scope of CRL in the future.

\section*{Acknowledgements}
This work is partially supported by the National Key Research and Development Program of China (under Grant 2018AAA0103300), the NSF of China (under Grants 61925208, 62002338, 62102399, U19B2019, 61732020), Beijing Academy of Artificial Intelligence (BAAI), CAS Project for Young Scientists in Basic Research (YSBR-029), Youth Innovation Promotion Association CAS and Xplore Prize.

\bibliography{aaai23}

\end{document}